\documentclass{article}

\usepackage[preprint]{neurips_2025}

\usepackage[utf8]{inputenc} 
\usepackage[T1]{fontenc}    
\usepackage{hyperref}       
\usepackage{url}            
\usepackage{booktabs}       
\usepackage{amsfonts}       
\usepackage{nicefrac}       
\usepackage{microtype}      
\usepackage{xcolor}         
\usepackage{multicol}
\usepackage{multirow}
\usepackage{amsmath}
\usepackage{natbib}
\usepackage{algorithm}
\usepackage{algorithmic}
\usepackage{cleveref}
\usepackage{graphicx}
\usepackage{colortbl}
\usepackage{wrapfig}
\usepackage{subfigure}

\graphicspath{{figures}}
\BeforeBeginEnvironment{wrapfigure}{\setlength{\intextsep}{0pt}}

\title{Temporal Saliency-Guided Distillation: A Scalable Framework for Distilling Video Datasets}

\author{
 Xulin Gu$^{1}$\thanks{Equal Contribution.} \quad
 Xinhao Zhong$^{1 *}$\quad Zhixing Wei$^{1}$\quad  Yimin Zhou$^{3}$\quad Shuoyang Sun$^{1}$\quad \\
 \textbf{Bin Chen}$^{1, 2}$\thanks{Corresponding Author.}\quad 
 \textbf{Hongpeng Wang}$^{1}$\quad   \quad  \textbf{Yuan Luo}$^4$\\
$^1$Harbin Institute of Technology, Shenzhen \quad $^2$Peng Cheng Laboratory \\ 
$^3$Tsinghua Shenzhen International Graduate School \quad $^4$Shanghai Jiao Tong University\\
    \footnotesize{\texttt{guxulin@stu.hit.edu.cn,}}
    \footnotesize{\texttt{xh021213@gmail.com,}}
    \footnotesize{\texttt{220111031@stu.hit.edu.cn,}} \\
    \footnotesize{\texttt{zhou-ym24@mails.tsinghua.edu.cn,}}
    \footnotesize{\texttt{24s151152@stu.hit.edu.cn,}} \\
    \footnotesize{\texttt{chenbin2021@hit.edu.cn,}}
    \footnotesize{\texttt{wanghp@hit.edu.cn,}}
    \footnotesize{\texttt{luoyuan@cs.sjtu.edu.cn,}}
    \vspace{-2em}
}

\begin{document}

\maketitle

\begin{abstract}
Dataset distillation (DD) has emerged as a powerful paradigm for dataset compression, enabling the synthesis of compact surrogate datasets that approximate the training utility of large-scale ones. While significant progress has been achieved in distilling image datasets, extending DD to the video domain remains challenging due to the high dimensionality and temporal complexity inherent in video data. Existing video distillation (VD) methods often suffer from excessive computational costs and struggle to preserve temporal dynamics, as naïve extensions of image-based approaches typically lead to degraded performance. In this paper, we propose a novel uni-level video dataset distillation framework that directly optimizes synthetic videos with respect to a pre-trained model. To address temporal redundancy and enhance motion preservation, we introduce a temporal saliency-guided filtering mechanism that leverages inter-frame differences to guide the distillation process, encouraging the retention of informative temporal cues while suppressing frame-level redundancy. Extensive experiments on standard video benchmarks demonstrate that our method achieves state-of-the-art performance, bridging the gap between real and distilled video data and offering a scalable solution for video dataset compression.
\end{abstract}

\section{Introduction}
\begin{wrapfigure}{r}{0.55\linewidth}
  \centering
  \vspace{-1em}
  \includegraphics[width = \linewidth]{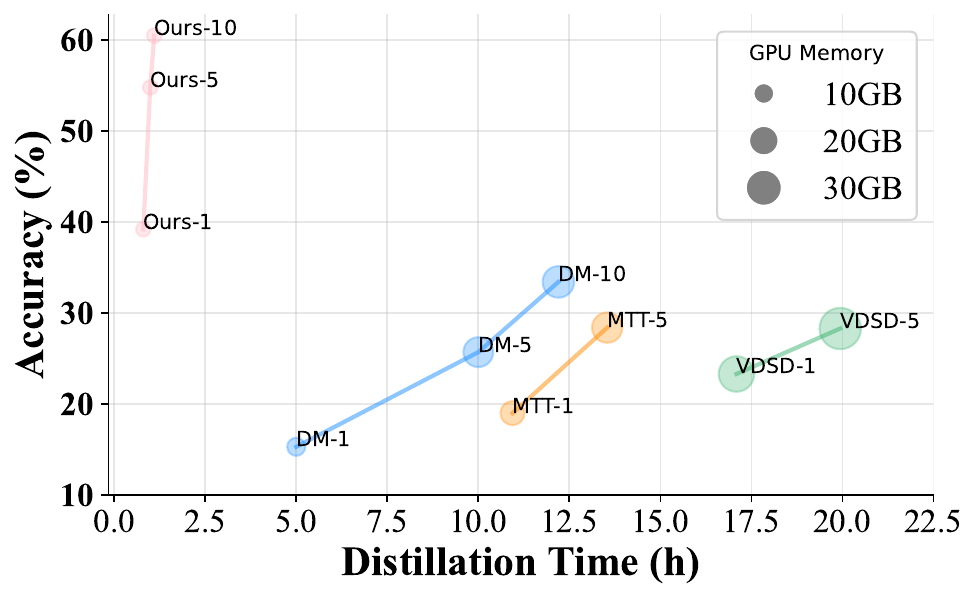}
  \vspace{-2em}
  \caption{Comparison of test accuracy and distillation cost between different methods across all the IPC settings. Our method exhibits superior performance.}
  \label{fig:intro}
\end{wrapfigure}
In recent years, video data has accounted for over 70\% of global internet traffic, and this proportion continues to grow steadily \cite{barnett2018cisco}. The explosive increase in online video content has driven a strong demand for automatic video understanding, which has in turn led to remarkable progress in video action recognition using neural networks. However, the impressive performance of these models often comes at the cost of high computational expense \cite{wang2021adaptive}. The sheer volume of video data, along with its significantly larger size compared to images, imposes substantial training costs and computational burdens on the models. Dataset distillation has become a promising technique to address this challenge, aiming to compress large-scale datasets into compact synthetic datasets while preserving as much task-specific information as possible.

However, existing dataset distillation methods still primarily focus on the image domain and are difficult to directly extend to videos. Existing methods can be broadly classified into two categories: optimization-based and training-free methods. 
Traditional optimization-based methods~\cite{wang2018dataset,zhao2021dataset,nguyen2021dataset,zhao2021datasetdm,cazenavette2022dataset,zhong2024going,sajedi2023datadam,cazenavette2023generalizing,zhong2024hierarchical} typically employ a bi-level optimization framework, wherein the synthetic dataset is optimized in conjunction with the training of a neural network that functions as a task-specific information extractor. Another category of optimization-based methods, decoupled distillation methods~\cite{yin2024squeeze,shao2024generalized,yin2024dataset,du2024diversity,shao2024elucidating}, optimize synthetic datasets without repeatedly training random sampled neural networks to further reduce cost. Although these methods often achieve strong performance under high compression ratios, the need of optimize all the pixels with a synthetic dataset introduces considerable time overhead and memory consumption. In contrast, training-free methods~\cite{sun2024diversity,su2024d,gu2024efficient,cheninfluence} commonly utilize the priors from pre-trained classifiers or generative models to directly synthesize datasets, avoiding pixel-level optimization of individual images and thereby significantly reducing computational costs. However, due to their limited capacity to remove redundant information, such methods generally result in suboptimal performance.

Although effective for image datasets, these two types of methods encounter significant challenges when applied to video datasets. Videos comprise sequences of frames, which significantly amplifies the computational cost of traditional optimization-based approaches, making them impractical for high-resolution or large-scale video distillation scenarios. Moreover, the temporal dimension inherent in video data necessitates that the distillation process account for inter-frame relationships to preserve temporal coherence and motion dynamics. In contrast, training-free methods are limited by the current capabilities of video generation models. While image generation techniques have reached a level of maturity that allows for the synthesis of high-quality images, video generation models still struggle to produce high-fidelity sequences suitable for action recognition tasks, often suffering from fragmented or incoherent temporal dynamics. Additionally, the presence of temporal dynamics greatly increases the difficulty of selecting informative frames, further contributing to the suboptimal performance of these approaches.

To address these challenges, VDSD~\cite{wang2024dancing} introduces a two-stage video distillation paradigm. In the first stage, it randomly selects frames from each video and performs standard image-level distillation. In the second stage, an interpolation model is used to reconstruct videos, followed by image distillation applied to each individual frame. Although this method achieves certain performance, the two stages are inherently conflicting and fail to effectively incorporate temporal information. As an extention of VDSD, IDTD~\cite{zhao2024video} selects frames with maximal differences to enhance frame diversity in the first stage and randomly sampling frames for interpolation in the second stage, thereby increasing video variability. While IDTD demonstrates improvements over VDSD, it maintains the same distillation pipeline and still overlooks essential temporal dependencies. Furthermore, the two-stage framework introduces considerable time overhead while offering only marginal performance gains.

However, extending image dataset distillation methods to videos and preserving temporal dynamics during the distillation process is a non-trivial task. First, modeling temporal dynamics is itself a difficult task. While using optical flow as a representation is a straightforward solution, it introduces considerable storage and computational overhead, necessitating more compact and efficient alternatives. Second, performing data augmentation on video without disrupting its temporal coherence is another major concern. Differentiable data augmentation has proven effective in enhancing the diversity of synthetic data in dataset distillation. However, naively applying image-level augmentation techniques to videos can degrade them into a sequence of unrelated frames, ultimately impairing performance. Therefore, designing augmentation strategies that respect the temporal structure of videos is crucial for effective video dataset distillation.

In this paper, we propose a novel framework for video dataset distillation. By simplifying the conventional bi-level optimization framework, we design an efficient uni-level optimization-based distillation process. Furthermore, we introduce the \textbf{T}emporal \textbf{S}aliency-\textbf{G}uided \textbf{F}ilter (TSGF) to preserve and enhance temporal information during the distillation process. Specifically, we first train a standard video classification model to encode video information into a pre-trained classifier. Subsequently, we align the distribution of the synthetic dataset with the model's internal statistics under the guidance of classification loss. To further reinforce temporal information, we apply TSGF, computed via inter-frame differences, as a constraint during optimization. In the post-evaluation phase, we propose a dynamic video augmentation strategy that enhances synthetic videos with TSGF to identify the key frames. By effectively capturing and reinforcing temporal characteristics throughout the overall distillation process, our method achieves significant performance improvements across various benchmark datasets. As illustrated in Figure \ref{fig:intro}, compared to existing image and video distillation methods, our approach significantly improves performance while substantially reducing both computational time and memory consumption, demonstrating the effectiveness of our framework.

The contributions of this paper can be summarized as follows:
\begin{itemize} 
    \item We propose an efficient uni-level video dataset distillation framework that is orthogonal to existing methods. It effectively addresses the excessive time consumption of prior approaches and reduces framework complexity. 
    \item We introduce TSGF, a temporal saliency-guided filter to guide the distillation process, along with a dynamic data augmentation technique. Both components enhance the temporal consistency and informativeness of the generated videos. 
    \item Extensive experiments across various video datasets and compression ratios demonstrate the effectiveness and efficiency of our method, significantly reducing training time for video action classification tasks. 
\end{itemize}

\section{Related Work}
\subsection{Image Dataset Distillation} 
Traditional optimization-based dataset distillation methods were initially formulated as meta-learning problems~\cite{wang2018dataset}, in which the objective is achieved by alternately updating the neural network and the distilled dataset. To mitigate the substantial computational cost caused by the unrolled computational graph inherent in meta-learning, DC~\cite{zhao2021dataset} proposed matching the gradients of the loss function with respect to model parameters on both real and synthetic datasets, thereby significantly reducing overhead via short-horizon gradient matching. MTT~\cite{cazenavette2022dataset} extended this idea by aligning the parameter trajectories of models trained on real and synthetic datasets, enabling long-horizon matching. Alternatively, DM~\cite{zhao2021dataset} focused on aligning feature distributions between datasets by treating the neural network solely as a feature extractor, further reducing computational costs.

In contrast to optimization-based methods, training-free dataset distillation approaches avoid pixel-level optimization of the synthetic dataset and instead aim to preserve high-level semantic information. RDED~\cite{sun2024diversity} and DDPS~\cite{zhong2024efficient} respectively leverage a pre-trained classifier and a diffusion model to directly select the most class-representative images from the original dataset. Other methods such as Minimax~\cite{gu2024efficient}, D$^4$M~\cite{su2024d}, and IGD~\cite{cheninfluence} fine-tune components of diffusion models~\cite{song2020denoising} to directly generate synthetic images for the target dataset. While these approaches demonstrate promising performance on large-scale datasets~\cite{deng2009imagenet}, they tend to suffer under extreme compression ratios due to a lack of optimization, often resulting in synthetic datasets with redundant information and suboptimal performance.

Distinct from the aforementioned methods, decoupled dataset distillation methods~\cite{yin2024squeeze,yin2024dataset,shao2024elucidating,shao2024generalized,du2024diversity}, such as SRe$^2$L~\cite{yin2024squeeze}, first compress the information of the full dataset into a target model, and then generate synthetic datasets by aligning their statistical properties (e.g., batch normalization statistics) and task-specific loss with those of the model. However, a key limitation of existing work~\cite{chen2024large} that directly applies decoupled methods to video datasets lies in the treatment of frames from the same video as independent and unrelated samples, resulting in redundant optimization and neglect of temporal structure.

\subsection{Video Dataset Distillation} 
Compared to image data, video data incorporates an additional temporal dimension, which significantly increases the complexity of the distillation process. Currently, research on video dataset distillation remains in its early stages. VDSD~\cite{wang2024dancing} is the first work in this field, introducing a two-stage paradigm that disentangles the static and dynamic components of video data. In the static distillation stage, VDSD compresses pixel-level information into a single frame sampled from the video, aiming to align spatial features between the original and synthetic data. In the dynamic distillation stage, the distilled image is interpolated into a video, and semantic motion information is utilized to achieve temporal alignment between the original and synthetic sequences.

IDTD~\cite{zhao2024video} builds upon this two-stage paradigm and achieves further progress by incorporating two key modules. The Information Diversification module augments the distilled data into multiple feature segments to enhance diversity, while the Temporal Densification module aggregates these segments into complete video clips to capture richer temporal dynamics. These components improve the algorithm’s capacity to compress temporal features, leading to better performance.

Despite these advances, existing video dataset distillation methods often suffer from conflicting optimization objectives and considerable computational overhead. In contrast to previous approaches, we utilize a pre-trained classifier to guide the generation of synthetic video datasets, effectively reducing computational cost while preserving temporal coherence to the greatest extent possible.

\section{Method}

\subsection{Preliminary}
The objective of dataset distillation is to compress a large-scale training set $\mathcal{T} = \{(\mathbf{x}_\mathcal{T}^i, y_\mathcal{T}^i)\}_{i=1}^\mathcal{|T|}$ into a distilled dataset $\mathcal{S} = \{(\mathbf{x}_\mathcal{S}^i, y_\mathcal{S}^i)\}_{i=1}^\mathcal{|S|}$ $\mathcal{(|S| << |T|)}$, while preserving the training accuracy as much as possible. The learning objective on $\mathcal{S}$ can be formulated as follow:
\begin{equation}
    \theta_{\mathcal{S}} = \arg\min\limits_{\theta}\mathbb{E}_{(\mathbf{x}_{\mathcal{\mathcal{S}}}, y_{\mathcal{S}}) \in \mathcal{S}}[l(\phi_{\theta_{\mathcal{S}}}(\mathbf{x}_\mathcal{S}), y_\mathcal{S})],
\end{equation}
where $l(\cdot, \cdot)$ denotes the typical loss function (e.g., cross-entropy loss), and $\phi_{\theta_{\mathcal{S}}}$ represents the neural network with parameter $\theta_\mathcal{S}$. The primary objective of the dataset distillation task is to generate synthetic data aimed at attaining a specific or minimal performance disparity on the original validation data when the same models are trained on the synthetic data and the original dataset, respectively. Thus, we aim to optimize $\mathcal{S}$ as follow:
\begin{equation}
    \arg \min \limits_{\mathcal{S}, |\mathcal{S}|}(\sup\{|l(\phi_{\theta_{\mathcal{T}}}(\mathbf{x}_{val}), y_{val}) - l(\phi_{\theta_{\mathcal{S}}}(\mathbf{x}_{val}), y_{val})|\}_{(\mathbf{x}_{val}, y_{val}) \in \mathcal{T}}),
\end{equation}
where $(\mathbf{x}_{val}, y_{val})$ denotes the image and label pair samppled from the validation set of $\mathcal{T}$. For a typical dataset distillation process, the synthetic dataset $\mathcal{S}$ is first initialized using random noise or randomly sampled instances. Then, the distillation loss is computed, and $\mathcal{S}$ is optimized based on the loss. However, for video datasets, each video consists of multiple frames and can be regarded as image data with a large batch size in terms of computational cost. Consequently, previous image-based distillation methods are difficult to apply due to the substantial overhead of video datasets.

\begin{algorithm}[t]
    \caption{Saliency-Guided Video Distillation}
    \begin{algorithmic}
        \REQUIRE Original dataset $\mathcal{T}$, distillation iterations $K$, learning rate $\eta$
        \STATE Train $\theta_\mathcal{T}$ on $\mathcal{T}$: $\theta_\mathcal{T} = \arg\min\limits_{\theta_\mathcal{T}} \mathcal{L}_{ce}(\phi_{\theta_\mathcal{T}}(\mathbf{x}_{T}), y_{T})$
        \FOR{$k\xleftarrow{} 0$ to $K - 1$}
        \STATE Calculate regularization loss $\mathcal{L}_{reg}$ using Eq. (\ref{eq:reg})
        \STATE Calculate distillation loss $\mathcal{L} = \mathcal{L}_{ce}(\phi_{\theta_\mathcal{T}}(\mathbf{x}_S), y) + \mathcal{L}_{reg}$
        \STATE Compute mask $M$ using Eq. (\ref{eq:mask}), where compute temporal saliency vector $s$ using Eq. (\ref{eq:diff}) and Eq. (\ref{eq:smooth})
        \STATE Update $\mathcal{S} \xleftarrow{} \mathcal{S} - \eta M \nabla_\mathcal{S}\mathcal{L}$
        \ENDFOR
        \STATE Compute temporal saliency vector $s$ using Eq. (\ref{eq:diff}) and Eq. (\ref{eq:smooth})
        \STATE Apply temporally guided video augmentation: $\mathbf{x}_S = D(\mathbf{x}_S)$
        \STATE Recalibrate video data labels: $y_S = \phi_{\theta_{\mathcal{T}}}(\mathbf{x}_S)$ 
        \ENSURE Distilled dataset $\mathcal{S} = (\mathbf{x}_\mathcal{S}, y_\mathcal{S})$
    \end{algorithmic}
    \label{alg}
\end{algorithm}

\subsection{Uni-Level VD Framework}
\begin{figure}
  \centering
  \includegraphics[width = 0.95\linewidth]{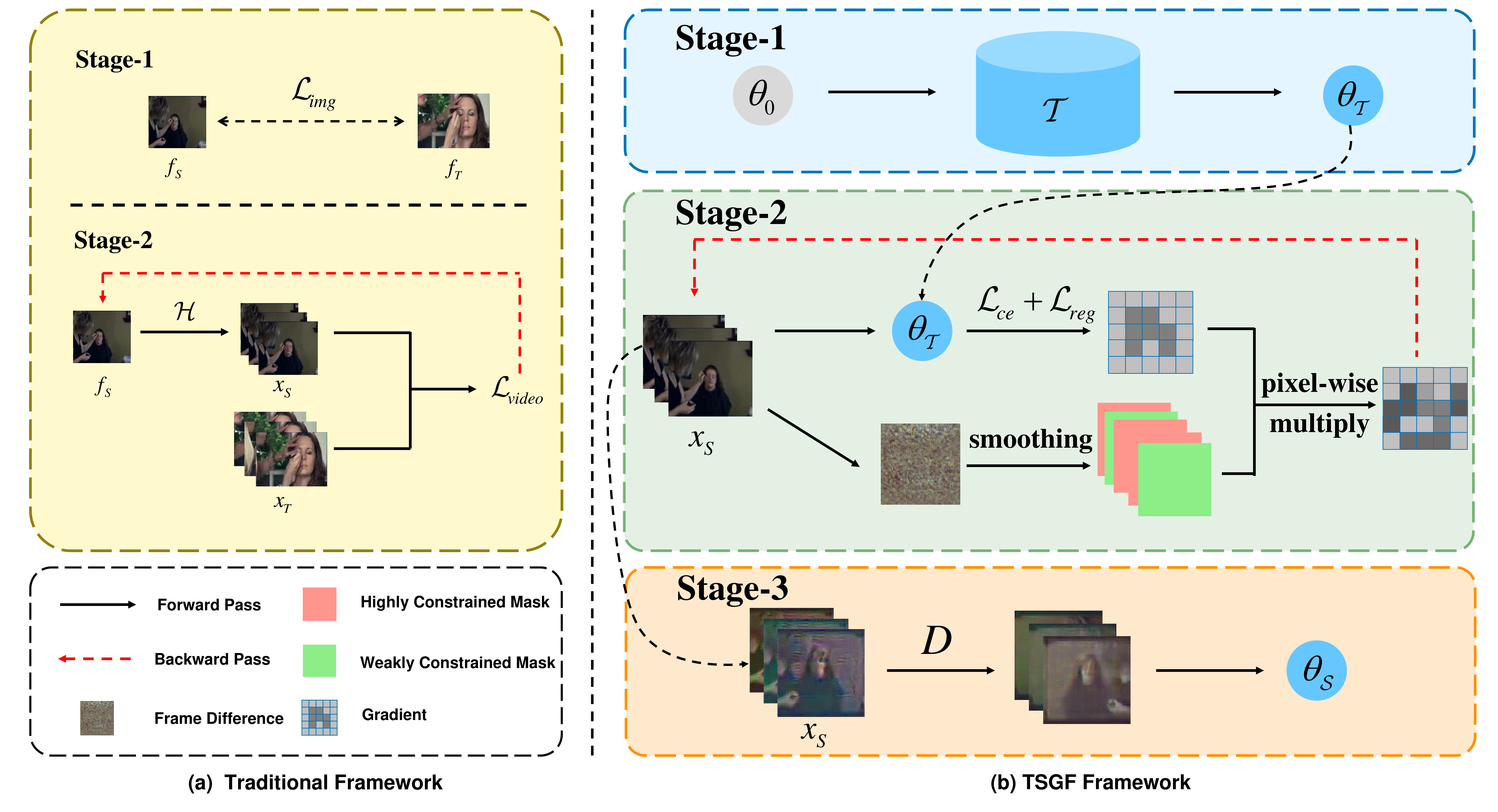}
  \caption{Comparison between our TSGF framework and the traditional two-stage video distillation paradigm. While both stages of the traditional approach primarily rely on pixel-level information, our unified framework effectively distills temporal information through a three-stage process. TSGF comprises two key components: TSGF$_O$and TSGF$_A$. During the optimization stage, TSGF$_O$ constrains the optimization by computing inter-frame differences. In the evaluation stage, TSGF$_A$ guides the data augmentation process to preserve temporal dynamics.}
  \label{fig:pipeline}
\end{figure}

Distinct from existing two-stage video distillation frameworks, we propose a novel uni-level video dataset distillation paradigm that enables efficient synthetic data generation. By decoupling model training from synthetic data optimization, our approach effectively addresses the substantial intra-batch computational cost encountered by traditional image-based distillation methods when applied to video data.

To capture the dataset distribution, we first train a video action recognition model, compressing the informative content of the original dataset into the model. The training process is formulated as:
\begin{equation}
\theta_\mathcal{T} = \arg\min\limits_{\theta_\mathcal{T}} \mathcal{L}_{ce}(\phi_{\theta_\mathcal{T}}(\mathbf{x}_{\mathcal{T}}), y_{\mathcal{T}}),
\end{equation}
where $\theta_\mathcal{T}$ denotes the parameters of the pre-trained model, and $\mathcal{L}_{ce}$ is the cross-entropy loss. Subsequently, we optimize the synthetic video samples using class-discriminative and statistical information extracted from the pre-trained model, following a process similar to~\cite{yin2020dreaming}:
\begin{equation}
\mathbf{x}_{\mathcal{S}} = \arg\min\limits_{\mathbf{x}_\mathcal{S}} \mathcal{L}_{ce}(\phi_{\theta_\mathcal{T}}(\mathbf{x}_\mathcal{S}), y) + \mathcal{L}_{reg}
\end{equation}
where $y$ is the one-hot class label assigned to $\mathbf{x}_\mathcal{S}$, and $\mathcal{L}_{reg}$ is a regularization term used to align the statistical distribution of synthetic data with the pre-trained model. The regularization loss is computed as follows:
\begin{equation}\label{eq:reg}
\mathcal{L}_{reg} = \sum_l \left(||\mu_l(\mathbf{x}_S) - RM_l||_2 + ||\sigma_l^2(\mathbf{x}_S) - RV_l||_2\right)
\end{equation}
where $l$ denotes the index of the batch normalization (BN) layer, $\mu_l(\mathbf{x}_S)$ and $\sigma_l^2(\mathbf{x}_S)$ represent the mean and variance of the activations in the $l$-th BN layer, and $RM_l$, $RV_l$ are the corresponding running mean and running variance.

\subsection{Temporal Saliency-Guided Filter} 

Naively treating video frames as discrete and independent images and applying uniform optimization across the entire video often leads to the loss of motion-related information in the temporal dimension, resulting in suboptimal performance~\cite{chen2024large}. To address this limitation, we propose a \textbf{T}emporal \textbf{S}aliency-\textbf{G}uided \textbf{F}ilter (TSGF), which comprises temporally guided video optimization and temporally guided video augmentation. This filter guides the update of video data by preserving and enhancing temporal coherence, thereby improving the overall quality of the distilled videos.

\textbf{Temporally Guided Video Optimization. }To preserve motion-related semantic information during the optimization process, it is essential to impose constraints that prevent over-optimization, which could otherwise diminish temporal dynamics. To this end, we compute the temporal saliency of each frame and assign adaptive optimization magnitudes accordingly. Specifically, we calculate the inter-frame difference for each frame using the following formulation:
\begin{equation}\label{eq:diff}
    d_i = \frac{|f_{i + 1} - f_i| + |f_i - f_{i - 1}|}{2}
\end{equation}
where $f_i$ denotes the $i$-th frame, and $d_i$ represents the inter-frame difference of the $i$-th frame. Inter-frame differences are often used as a metric to determine key frames in a video, and the magnitude of these differences can indicate the importance of each frame to some extent. However, the raw inter-frame difference cannot be directly used to measure the temporal saliency of the video frames because of the local jitter. To take advantage of the locality inherent in the video and eliminate the effects of local jitter, we apply smoothing to the raw inter-frame differences as follows:
\begin{equation}\label{eq:smooth}
    s_i = \alpha_0 * d_i + \alpha_{1} * d_{i - 1} + \cdots + \alpha_{k} * d_{i - k},
\end{equation}
where $s_i$ denotes the temporal saliency of the $i$-th frame, $k$ represents the window length, and $\alpha_k$ is the smoothing weight for previous frames, which is determined by the window function. To preserve the temporal information of the video, we compute the TSGF based on $s_{i}$ of each frame. Specifically, the greater the importance of a frame, the smaller its optimization degree will be. The calculation formula is as follows:
\begin{equation}\label{eq:mask}
    M = \frac{\max(\epsilon - s, 0)}{\max(s) - \min(s)},
\end{equation}
where $M$ denotes the optimization mask, $s$ represents the temporal saliency vector of the video, and $\epsilon$ indicates the upper bound of temporal saliency. The resulting mask is ultimately used to guide the optimization process of each frame in the video:
\begin{equation}
    \hat{g} = M \cdot g,
\end{equation}
where $g$ denotes the original gradient obtained from the distillation loss, and $\hat{g}$ is the normalized gradient ultimately used for video data optimization.

\textbf{Temporally Guided Video Augmentation.} To further enhance the diversity of synthetic datasets while preserving teporal information, we apply a dynamic video data augmentation during post-evaluation phase. To prevent the loss of key frame information during augmentation, we also utilize TSGF as guidance to selectively perform data augmentation. The augmented videos are then used for the final model training.

Common image augmentation methods such as MixUp~\cite{zhang2017mixup} and CutMix~\cite{yun2019cutmix} are not directly applicable to videos, as they can disrupt the inherent temporal structure. To address this issue, we first compute $s_{i}$ of each video frame using the Eq. (\ref{eq:smooth}). Then, we apply VideoMix~\cite{yun2020videomix} augmentation at the same positions across non-key frames, guided by their temporal saliency scores:
\begin{equation}
    D(f_i) = 
    \begin{cases}
        A(f_i) & s_i \leq \epsilon \\
        f_i    & else
    \end{cases},
\end{equation}
where $D(\cdot)$ represents the dynamic augmentation process applied to $f_i$, while $A(\cdot)$ denotes the actual data augmentation applied to $f_i$. Our distillation process is summarized in Algorithm \ref{alg}.

\begin{table}[htbp]
    \centering
    \vspace{0.5cm}
    \caption{Comparison of top-1 accuracy with existing methods on small-scale datasets. Our method achieve significant performance improvements across various settings.}
    \begin{tabular}{l|lcccc}
        \toprule
        \multicolumn{2}{c}{Dataset}  & \multicolumn{2}{c}{MiniUCF} & \multicolumn{2}{c}{HMDB51}  \\
        \midrule
        \multicolumn{2}{c}{IPC} & 1 & 5 & 1 & 5 \\
        \midrule
         & Random & $9.9\pm0.8$ & $22.9\pm1.1$ & $4.6\pm0.5$ & $6.6\pm0.7$ \\
         Coreset Selection & Herding & $12.7\pm1.6$ & $25.8\pm0.3$ & $3.8\pm0.2$ & $8.5\pm0.4$ \\
        & K-Center & $11.5\pm0.7$ & $23.0\pm1.3$ & $3.1\pm0.1$ & $5.2\pm0.3$ \\
        \midrule
         & DM & $15.3\pm1.1$ & $25.7\pm0.2$ & $6.1\pm0.2$ & $8.0\pm0.2$ \\
        & MTT & $19.0\pm0.1$ & $28.4\pm0.7$ & $6.6\pm0.5$ & $8.4\pm0.6$ \\
         & FRePo & $20.3\pm0.5$ & $30.2\pm1.7$ & $7.2\pm0.8$ & $9.6\pm0.7$ \\
        \multirow{2}{*}{Dataset Distillation}& DM+VDSD & $17.5\pm0.1$ & $27.2\pm0.4$ & $6.0\pm0.9$ & $8.2\pm0.4$ \\
        & MTT+VDSD & $23.3\pm0.6$ & $28.3\pm0.0$ & $6.5\pm0.4$ & $8.9\pm0.1$ \\
        & FRePo+VDSD & $22.0\pm1.0$ & $31.2\pm0.7$ & $8.6\pm0.1$ & $10.3\pm0.6$ \\
        & IDTD & $22.5\pm0.1$ & $33.3\pm0.5$ & $9.5\pm0.3$ & $16.2\pm0.9$ \\
        \rowcolor{gray!20}
        & Ours & $\mathbf{39.2\pm0.7}$ & $\mathbf{54.8\pm0.5}$ & $\mathbf{13.9\pm0.9}$ & $\mathbf{20.2\pm0.4}$ \\
        \bottomrule
    \end{tabular}
    \label{tab:main}
\end{table}

\section{Experiments}
To verify the efficiency of our proposed method, we conduct experiments following two-steps paradigm, as the standard evaluation procedure for dataset distillation. In the first step, the original dataset is compressed to obtain a synthetic dataset. In the second step, a standard model training process is conducted using the synthetic dataset, and performance is evaluated on the original test set.

\subsection{Datasets}
In this study, we conduct experiments on several widely-used video benchmark datasets, including UCF101\cite{soomro2012dataset}, HMDB51\cite{kuehne2011hmdb}, Kinetics-400\cite{carreira2017quo}, and Something-Something V2\cite{goyal2017something}. To ensure fair comparison with previous work~\cite{wang2024dancing,zhao2024video}, we adopt the light-weight version of UCF101, namely MiniUCF.

UCF101 is a widely used benchmark dataset for human action recognition, containing 13,320 unconstrained video clips from 101 action categories. The dataset covers a broad range of activities, including sports, human-object interactions, and body movements. HMDB51 consists of 6,766 video clips spanning 51 human action classes. The dataset features high intra-class variation and challenging scenarios due to its sources from movies and public databases. Kinetics-400 is a large-scale video dataset introduced by DeepMind, comprising around 240,000 training videos and 400 diverse action categories. Each video is approximately 10 seconds long and sourced from YouTube, providing rich variation in scene context and motion patterns for training deep video understanding models. Something-Something V2 focuses on fine-grained temporal reasoning and object interactions, with 220,847 video clips across 174 action categories. Unlike conventional datasets, it emphasizes abstract action concepts.

\subsection{Implementation Details}
To ensure fair comparison, all experimental settings are aligned with those used in VDSD~\cite{wang2024dancing} and IDTD~\cite{zhao2024video}. For the MiniUCF and HMDB51 datasets, we follow the dataset splits provided by VDSD, and each video is sampled to 16 frames with a resolution of 112×112. For the Kinetics-400 and SSv2 datasets, each video is sampled to 8 frames with a resolution of 64×64. The four datasets are grouped into two categories for experimentation: MiniUCF and HMDB51 are treated as light-weight datasets, and we report top-1 accuracy; Kinetics-400 and SSv2 are considered large-scale datasets, and we report top-5 accuracy. For the proxy model, we adopt MiniC3D\cite{wang2024dancing}, a light-weight version of the C3D\cite{tran2015learning} model obtained through architectural simplifications.

\begin{table}[tbp]
    \centering
    \caption{Comparison of top-5 accuracy with existing methods on large-scale datasets. $^{\dagger}$ denotes the top-5 accuracy of our teacher models trained on the full dataset is lower than the baselines due to different hyper-parameters settings.}
    \begin{tabular}{lcccc}
        \toprule
        Dataset & \multicolumn{2}{c}{Kinetics-400} & \multicolumn{2}{c}{SSv2}  \\
        \midrule
        IPC & 1 & 5 & 1 & 5 \\
        \midrule
        Random & $3.0\pm0.1$ & $5.6\pm0.0$ & $3.3\pm0.1$ & $3.9\pm0.1$ \\
        DM & $6.3\pm0.0$ & $9.1\pm0.9$ & $3.6\pm0.0$ & $4.1\pm0.0$ \\
        MTT & $3.8\pm0.2$ & $9.1\pm0.3$ & $3.9\pm0.1$ & $6.3\pm0.3$ \\
        DM+VDSD & $6.3\pm0.2$ & $7.0\pm0.1$ & $4.0\pm0.1$ & $3.8\pm0.1$ \\
        MTT+VDSD & $6.3\pm0.1$ & $11.5\pm0.5$ & $5.5\pm0.1$ & $8.3\pm0.1$ \\
        IDTD & $6.1\pm0.1$ & $12.1\pm0.2$ & $3.9\pm0.1$ & $9.5\pm0.3$ \\
        \rowcolor{gray!20}
        Ours & $\mathbf{6.5\pm0.2}^{\dagger}$ & $\mathbf{13.4\pm0.3}^{\dagger}$ & $\mathbf{10.4\pm0.1}$ & $\mathbf{15.2\pm0.2}$ \\
        \bottomrule
    \end{tabular}
    \label{tab:large}
\end{table}

\subsection{Main Results}
We present the experimental results on light-weight and large-scale datasets in Table \ref{tab:main} and Table \ref{tab:large}, respectively, demonstrating the superiority of our method across various datasets and IPC (image per class) settings. The methods evaluated include coreset selection methods, image dataset distillation methods, existing video dataset distillation methods and our method.

\textbf{Light-weight Datasets.} On the MiniUCF and HMDB51 datasets, our method achieves significantly better performance compared to existing approaches. Specifically, under the IPC=5 setting on MiniUCF, our method attains an accuracy of 54.8\%, yielding a 21.5\% improvement over previous methods. On HMDB51 with IPC=5, our method achieves 20.2\% accuracy, outperforming existing methods by 4\%. Although the performance gain on HMDB51 appears smaller compared to MiniUCF, it is important to note that the full dataset accuracy on HMDB51 is only 28.6\%. The accuracy achieved by our distilled dataset is already approaching that of the full dataset, demonstrating the effectiveness of our method. Coreset selection and image dataset distillation methods generally exhibit inferior performance, highlighting the limitations of image-based algorithms in effectively leveraging temporal information inherent in video data. These methods, when directly transferred to the video domain, can only achieve suboptimal results. 

Existing video distillation algorithms address this issue by incorporating partial temporal cues, leading to moderate performance gains; however, their improvements remain limited, as their effectiveness still primarily relies on spatial information, with minimal exploitation of temporal dynamics. Our proposed distillation algorithm achieves consistent performance improvements across all experimental settings, validating the effectiveness of the proposed temporal saliency-guided filter.

\textbf{Large-scale Datasets.} Under the IPC=5 setting on the SSv2 dataset, our method achieves an accuracy of 15.2\%, resulting in a 5.7\% performance gain over existing approaches. On Kinetics-400, our method also surpasses the baseline, even though the model used achieves only 22.4\% accuracy when trained on the full dataset. With an IPC=5 setting, the compression ratio on both datasets falls below 1\%, making the distillation task significantly more challenging than on light-weight datasets. Consequently, the overall performance of existing methods remains relatively low, with a substantial gap compared to models trained on the full datasets. This highlights the considerable room for improvement in video dataset distillation on large-scale datasets. Despite the increased difficulty, our method consistently outperforms existing approaches across all experimental settings.

\begin{figure}
  \centering
  \includegraphics[width = \textwidth]{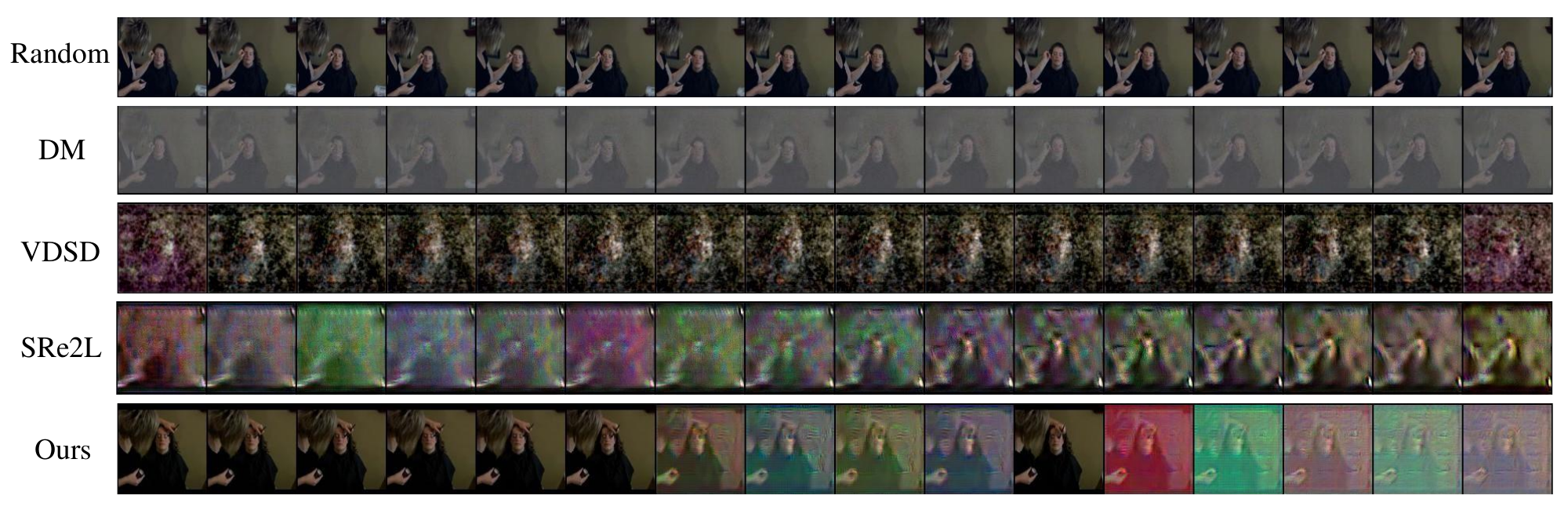}
  \vspace{-2em}
  \caption{Experimental results on MiniUCF under IPC=1 with varying numbers of frames.}
  \label{fig:visualization}
\end{figure}

\subsection{Visualization}

To visually demonstrate the effectiveness of our method in utilizing temporal information, we conduct a qualitative analysis of the distilled results under the IPC=1 setting on the MiniUCF dataset, with visualizations for each method presented in Figure \ref{fig:visualization}. As shown in the results, the distilled samples generated by the DM method are visually similar to real images. In contrast, VDSD suffers from distortion due to its reliance on interpolation techniques, and SRe$^2$L’s matching strategy leads to homogenization across frames, resulting in the loss of temporal information. In comparison, our method enables dynamic optimization of individual frames, effectively preserving temporal dynamics while simultaneously compressing spatial information.

\subsection{Ablation Study}

\begin{wraptable}{r}{0.5\textwidth}
\centering
\caption{Ablation study on individual components. TSGF$_{O}$ and TSGF$_{A}$ denote the video optimization and the video augmentation based on TSGF, respectively.}
\resizebox{0.5\textwidth}{!}{
\begin{tabular}{cc}
\toprule
 Methods & Acc \\
\midrule
baseline & $40.5\pm0.3$ \\
baseline + TSGF$_{A}$ & $46.9\pm0.2$ \\
baseline + TSGF$_{O}$ & $51.7\pm0.6$ \\
\rowcolor{gray!20}
baseline + TSGF$_{O}$ +TSGF$_{A}$ & $\mathbf{54.8\pm0.5}$ \\
\bottomrule
 \end{tabular}}
\label{tab:ablation}
\end{wraptable}
\textbf{Effectiveness of Each Component.}
\quad To verify the effectiveness of each component in the distillation framework, we conduct ablation studies on the MiniUCF dataset under the IPC=5 setting. The detailed results are shown in Table \ref{tab:ablation}, where TSGF$_{O}$ denotes the temporal saliency-guided filter, and TSGF$_{A}$ refers to the temporal saliency-aware video augmentation strategy. The baseline refers to the three-stage video distillation framework without the inclusion of these two components. Experimental results demonstrate that both components contribute significantly to performance improvements.

\begin{wraptable}{r}{0.5\textwidth}
        \centering
        \caption{Cross-architecture experiments on MiniUCF with IPC=1. Our method achieve superior generalization ability. }
        \resizebox{0.5\textwidth}{!}{%
        \begin{tabular}{c|ccc}
            \toprule
             & \multicolumn{3}{c}{Evaluation Model} \\
             & ConvNet3D & CNN+GRU & CNN+LSTM \\
            \midrule
            Random & $9.9\pm0.8$ & $6.2\pm0.8$ & $6.5\pm0.3$ \\
            DM & $15.3\pm1.1$ & $9.9\pm0.7$ & $9.2\pm0.3$ \\
            MTT & $19.0\pm0.1$ & $8.4\pm0.5$ & $7.3\pm0.4$ \\
            DM+VDSD & $17.5\pm0.1$ & $12.0\pm0.7$ & $10.3\pm0.2$ \\
            MTT+VDSD & $23.3\pm0.6$ & $14.8\pm0.1$ & $13.4\pm0.2$ \\
            \rowcolor{gray!20}
            Ours & $\mathbf{39.2\pm0.7}$ & $\mathbf{17.8\pm0.4}$ & $\mathbf{16.5\pm0.3}$ \\
            \bottomrule
        \end{tabular}}
        \label{tab:generalization}
\end{wraptable}

\textbf{Cross Architecture Generalization.} We present the cross architecture generalization results in Table \ref{tab:generalization}, where all the evaluated model architectures are introduced in \cite{wang2024dancing}. Although the synthetic dataset is generated based on a pre-trained model, which limits the performance gain when transferring across architectures, our method consistently outperforms existing approaches across all architectures. This demonstrates the strong generalization ability of our method in cross-architecture scenarios.

\begin{wrapfigure}{r}{0.5\linewidth}
  \centering
  \includegraphics[width = \linewidth]{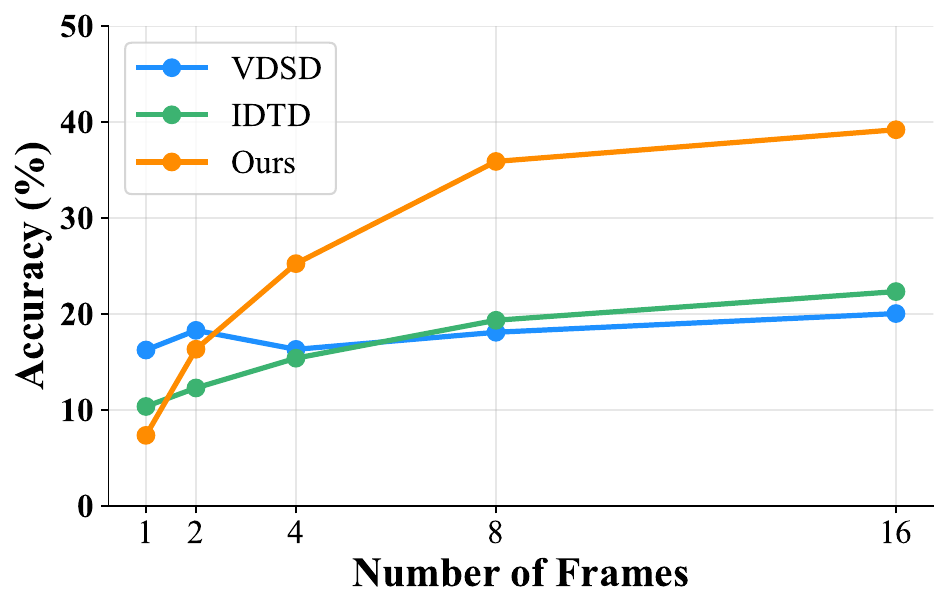}
  \vspace{-2em}
  \caption{Experimental results on MiniUCF under IPC=1 with varying numbers of frames.}
  \label{fig:frames}
  \vspace{-1em}
\end{wrapfigure}
\textbf{Number of Frames.} Figure \ref{fig:frames} presents the accuracy comparison between our method and the baseline under different numbers of frames. As shown, when the number of frames is small, the video data essentially degrades into image data, with minimal temporal information available. In this case, our method achieves relatively poor performance. However, as the number of frames increases and temporal information becomes more prominent, the performance of our method improves significantly, demonstrating its effectiveness in leveraging temporal cues in video data.

\section{Conclusion and Limitations}
In this study, we propose a unified video distillation framework that enables efficient video distillation based on a Temporal Saliency-Guided Filter. We first train the model using a standard training process, and then perform temporally guided video optimization to simultaneously compress spatial and motion-related information during the distillation process. To further enhance video diversity, we adopt temporally guided video augmentation to augment the video data. Extensive experiments conducted on datasets of varying scales and domains, demonstrating that our method provides a novel and effective paradigm for video dataset distillation. 

We recognize the limitations of our work from two perspectives. First, large-scale datasets remain a highly challenging scenario for our method, as there is still a considerable performance gap compared to models trained on the full dataset. Second, our approach currently lacks the incorporation of motion compensation information such as optical flow. Effectively integrating such information could further enhance the compression of temporal dynamics.

\bibliographystyle{plain}
\bibliography{reference}

\begin{thebibliography}{10}

\bibitem{barnett2018cisco}
Thomas Barnett, Shruti Jain, Usha Andra, and Taru Khurana.
\newblock Cisco visual networking index (vni) complete forecast update, 2017--2022.
\newblock {\em Americas/EMEAR Cisco Knowledge Network (CKN) Presentation}, 1(1), 2018.

\bibitem{carreira2017quo}
Joao Carreira and Andrew Zisserman.
\newblock Quo vadis, action recognition? a new model and the kinetics dataset.
\newblock In {\em proceedings of the IEEE Conference on Computer Vision and Pattern Recognition}, pages 6299--6308, 2017.

\bibitem{cazenavette2022dataset}
George Cazenavette, Tongzhou Wang, Antonio Torralba, Alexei~A Efros, and Jun-Yan Zhu.
\newblock Dataset distillation by matching training trajectories.
\newblock In {\em Proceedings of the IEEE/CVF Conference on Computer Vision and Pattern Recognition}, pages 4750--4759, 2022.

\bibitem{cazenavette2023generalizing}
George Cazenavette, Tongzhou Wang, Antonio Torralba, Alexei~A Efros, and Jun-Yan Zhu.
\newblock Generalizing dataset distillation via deep generative prior.
\newblock {\em arXiv preprint arXiv:2305.01649}, 2023.

\bibitem{cheninfluence}
Mingyang Chen, Jiawei Du, Bo~Huang, Yi~Wang, Xiaobo Zhang, and Wei Wang.
\newblock Influence-guided diffusion for dataset distillation.
\newblock In {\em The Thirteenth International Conference on Learning Representations}, 2025.

\bibitem{chen2024large}
Yang Chen, Sheng Guo, Bo~Zheng, and Limin Wang.
\newblock A large-scale study on video action dataset condensation.
\newblock {\em arXiv preprint arXiv:2412.21197}, 2024.

\bibitem{deng2009imagenet}
Jia Deng, Wei Dong, Richard Socher, Li-Jia Li, Kai Li, and Li~Fei-Fei.
\newblock Imagenet: A large-scale hierarchical image database.
\newblock In {\em 2009 IEEE conference on computer vision and pattern recognition}, pages 248--255. Ieee, 2009.

\bibitem{du2024diversity}
Jiawei Du, Xin Zhang, Juncheng Hu, Wenxin Huang, and Joey~Tianyi Zhou.
\newblock Diversity-driven synthesis: Enhancing dataset distillation through directed weight adjustment.
\newblock {\em arXiv preprint arXiv:2409.17612}, 2024.

\bibitem{goyal2017something}
Raghav Goyal, Samira Ebrahimi~Kahou, Vincent Michalski, Joanna Materzynska, Susanne Westphal, Heuna Kim, Valentin Haenel, Ingo Fruend, Peter Yianilos, Moritz Mueller-Freitag, et~al.
\newblock The" something something" video database for learning and evaluating visual common sense.
\newblock In {\em Proceedings of the IEEE international conference on computer vision}, pages 5842--5850, 2017.

\bibitem{gu2024efficient}
Jianyang Gu, Saeed Vahidian, Vyacheslav Kungurtsev, Haonan Wang, Wei Jiang, Yang You, and Yiran Chen.
\newblock Efficient dataset distillation via minimax diffusion.
\newblock In {\em Proceedings of the IEEE/CVF Conference on Computer Vision and Pattern Recognition}, pages 15793--15803, 2024.

\bibitem{kuehne2011hmdb}
Hildegard Kuehne, Hueihan Jhuang, Est{\'\i}baliz Garrote, Tomaso Poggio, and Thomas Serre.
\newblock Hmdb: a large video database for human motion recognition.
\newblock In {\em 2011 International conference on computer vision}, pages 2556--2563. IEEE, 2011.

\bibitem{nguyen2021dataset}
Timothy Nguyen, Roman Novak, Lechao Xiao, and Jaehoon Lee.
\newblock Dataset distillation with infinitely wide convolutional networks.
\newblock {\em Advances in Neural Information Processing Systems}, 34:5186--5198, 2021.

\bibitem{sajedi2023datadam}
Ahmad Sajedi, Samir Khaki, Ehsan Amjadian, Lucy~Z Liu, Yuri~A Lawryshyn, and Konstantinos~N Plataniotis.
\newblock Datadam: Efficient dataset distillation with attention matching.
\newblock In {\em Proceedings of the IEEE/CVF International Conference on Computer Vision}, pages 17097--17107, 2023.

\bibitem{shao2024generalized}
Shitong Shao, Zeyuan Yin, Muxin Zhou, Xindong Zhang, and Zhiqiang Shen.
\newblock Generalized large-scale data condensation via various backbone and statistical matching.
\newblock In {\em Proceedings of the IEEE/CVF Conference on Computer Vision and Pattern Recognition}, pages 16709--16718, 2024.

\bibitem{shao2024elucidating}
Shitong Shao, Zikai Zhou, Huanran Chen, and Zhiqiang Shen.
\newblock Elucidating the design space of dataset condensation.
\newblock {\em arXiv preprint arXiv:2404.13733}, 2024.

\bibitem{song2020denoising}
Jiaming Song, Chenlin Meng, and Stefano Ermon.
\newblock Denoising diffusion implicit models.
\newblock {\em ICLR}, 2021.

\bibitem{soomro2012dataset}
Khurram Soomro, Amir~Roshan Zamir, and Mubarak Shah.
\newblock A dataset of 101 human action classes from videos in the wild.
\newblock {\em Center for Research in Computer Vision}, 2(11):1--7, 2012.

\bibitem{su2024d}
Duo Su, Junjie Hou, Weizhi Gao, Yingjie Tian, and Bowen Tang.
\newblock D\^{} 4: Dataset distillation via disentangled diffusion model.
\newblock In {\em Proceedings of the IEEE/CVF Conference on Computer Vision and Pattern Recognition}, pages 5809--5818, 2024.

\bibitem{sun2024diversity}
Peng Sun, Bei Shi, Daiwei Yu, and Tao Lin.
\newblock On the diversity and realism of distilled dataset: An efficient dataset distillation paradigm.
\newblock In {\em Proceedings of the IEEE/CVF Conference on Computer Vision and Pattern Recognition}, pages 9390--9399, 2024.

\bibitem{tran2015learning}
Du~Tran, Lubomir Bourdev, Rob Fergus, Lorenzo Torresani, and Manohar Paluri.
\newblock Learning spatiotemporal features with 3d convolutional networks.
\newblock In {\em Proceedings of the IEEE international conference on computer vision}, pages 4489--4497, 2015.

\bibitem{wang2018dataset}
Tongzhou Wang, Jun-Yan Zhu, Antonio Torralba, and Alexei~A Efros.
\newblock Dataset distillation.
\newblock {\em arXiv preprint arXiv:1811.10959}, 2018.

\bibitem{wang2021adaptive}
Yulin Wang, Zhaoxi Chen, Haojun Jiang, Shiji Song, Yizeng Han, and Gao Huang.
\newblock Adaptive focus for efficient video recognition.
\newblock In {\em proceedings of the IEEE/CVF international conference on computer vision}, pages 16249--16258, 2021.

\bibitem{wang2024dancing}
Ziyu Wang, Yue Xu, Cewu Lu, and Yong-Lu Li.
\newblock Dancing with still images: video distillation via static-dynamic disentanglement.
\newblock In {\em Proceedings of the IEEE/CVF Conference on Computer Vision and Pattern Recognition}, pages 6296--6304, 2024.

\bibitem{yin2020dreaming}
Hongxu Yin, Pavlo Molchanov, Jose~M Alvarez, Zhizhong Li, Arun Mallya, Derek Hoiem, Niraj~K Jha, and Jan Kautz.
\newblock Dreaming to distill: Data-free knowledge transfer via deepinversion.
\newblock In {\em Proceedings of the IEEE/CVF conference on computer vision and pattern recognition}, pages 8715--8724, 2020.

\bibitem{yin2024dataset}
Zeyuan Yin and Zhiqiang Shen.
\newblock Dataset distillation via curriculum data synthesis in large data era.
\newblock {\em Transactions on Machine Learning Research}, 2024.

\bibitem{yin2024squeeze}
Zeyuan Yin, Eric Xing, and Zhiqiang Shen.
\newblock Squeeze, recover and relabel: Dataset condensation at imagenet scale from a new perspective.
\newblock {\em Advances in Neural Information Processing Systems}, 36, 2024.

\bibitem{yun2019cutmix}
Sangdoo Yun, Dongyoon Han, Seong~Joon Oh, Sanghyuk Chun, Junsuk Choe, and Youngjoon Yoo.
\newblock Cutmix: Regularization strategy to train strong classifiers with localizable features.
\newblock In {\em Proceedings of the IEEE/CVF international conference on computer vision}, pages 6023--6032, 2019.

\bibitem{yun2020videomix}
Sangdoo Yun, Seong~Joon Oh, Byeongho Heo, Dongyoon Han, and Jinhyung Kim.
\newblock Videomix: Rethinking data augmentation for video classification.
\newblock {\em arXiv preprint arXiv:2012.03457}, 2020.

\bibitem{zhang2017mixup}
Hongyi Zhang, Moustapha Cisse, Yann~N Dauphin, and David Lopez-Paz.
\newblock mixup: Beyond empirical risk minimization.
\newblock {\em arXiv preprint arXiv:1710.09412}, 2017.

\bibitem{zhao2021datasetdm}
Bo~Zhao and Hakan Bilen.
\newblock Dataset condensation with distribution matching.
\newblock {\em arXiv preprint arXiv:2110.04181}, 2021.

\bibitem{zhao2021dataset}
Bo~Zhao, Konda~Reddy Mopuri, and Hakan Bilen.
\newblock Dataset condensation with gradient matching.
\newblock {\em ICLR}, 1(2):3, 2021.

\bibitem{zhao2024video}
Yinjie Zhao, Heng Zhao, Bihan Wen, Yew-Soon Ong, and Joey~Tianyi Zhou.
\newblock Video set distillation: Information diversification and temporal densification.
\newblock {\em arXiv preprint arXiv:2412.00111}, 2024.

\bibitem{zhong2024going}
Xinhao Zhong, Bin Chen, Hao Fang, Xulin Gu, Shu-Tao Xia, and En-Hui Yang.
\newblock Going beyond feature similarity: Effective dataset distillation based on class-aware conditional mutual information.
\newblock {\em arXiv preprint arXiv:2412.09945}, 2024.

\bibitem{zhong2024hierarchical}
Xinhao Zhong, Hao Fang, Bin Chen, Xulin Gu, Tao Dai, Meikang Qiu, and Shu-Tao Xia.
\newblock Hierarchical features matter: A deep exploration of gan priors for improved dataset distillation.
\newblock {\em arXiv preprint arXiv:2406.05704}, 2024.

\bibitem{zhong2024efficient}
Xinhao Zhong, Shuoyang Sun, Xulin Gu, Zhaoyang Xu, Yaowei Wang, Jianlong Wu, and Bin Chen.
\newblock Efficient dataset distillation via diffusion-driven patch selection for improved generalization.
\newblock {\em arXiv preprint arXiv:2412.09959}, 2024.

\end{thebibliography}

\clearpage
\appendix

\section*{Technical Appendices and Supplementary Material}

\section{Ablation study on initialization.}
\begin{table}[htbp]
    \centering
    \begin{minipage}[t]{0.28\linewidth}
        \centering
        \caption{Results under various initialization strategies}
        \resizebox{\linewidth}{!}{
            \begin{tabular}{cc}
            \toprule
             Methods & Acc \\
            \midrule
            random init & $36.8\pm0.3$ \\
            real init & $54.8\pm0.5$ \\
            \bottomrule
            \end{tabular}
        }
        \label{tab:init}
    \end{minipage}
    \begin{minipage}[t]{0.28\linewidth}
        \centering
        \caption{Ablation study on data augmentation.}
        \resizebox{\linewidth}{!}{
            \begin{tabular}{cc}
            \toprule
             Methods & Acc \\
            \midrule
            image-based & $43.1\pm0.4$ \\
            TSGF$_A$ & $54.8\pm0.5$ \\
            \bottomrule
            \end{tabular}
        }
        \label{tab:aug}
    \end{minipage}
    \begin{minipage}[t]{0.38\linewidth}
        \centering
        \caption{Experimental results on static and dynamic group.}
        \resizebox{\linewidth}{!}{
            \begin{tabular}{cccc}
                \toprule
                IPC & Acc & S\_Acc & D\_Acc \\
                \midrule
                1 & $39.2\pm0.7$ & $35.5\pm1.0$ & $34.6\pm0.8$ \\
                5 & $54.8\pm0.5$ & $53.6\pm1.2$ & $58.5\pm0.7$ \\
                10 & $60.5\pm0.6$ & $57.1\pm0.9$ & $64.6\pm0.7$ \\
                \bottomrule
            \end{tabular}}
        \label{tab:dynamic}
    \end{minipage}
\end{table}

To evaluate the impact of initialization on the distillation process, we conduct an ablation study by comparing different initialization strategies for the synthetic dataset. Specifically, we compare random noise initialization with initialization from real video frames. Our results indicate that real initialization significantly outperforms noise-based initialization. This suggests that temporal priors embedded in real videos provide a more informative starting point for learning temporal dynamics.

\section{IPC Ablation}

We further examine the effect of instance per class (IPC) on the performance of the distilled dataset. Experiments are conducted on MiniUCF and HMDB51 with IPC values ranging from 1 to 20. As expected, the performance improves consistently with larger IPC values, highlighting the trade-off between data compactness and model accuracy. 

\begin{table}[htbp]
    \centering
    \caption{Ablation study on IPC.}
    \begin{tabular}{ccccc}
        \toprule
        IPC & 1 & 5 & 10 & 20 \\
        \midrule
        MiniUCF & $39.2\pm0.7$ & $54.8\pm0.5$ & $60.5\pm0.6$ & $62.8\pm0.2$ \\
        HMDB51 & $13.9\pm0.9$ & $20.2\pm0.4$ & $22.5\pm0.9$ & $26.3\pm0.7$ \\
        \bottomrule
    \end{tabular}
    \label{tab:ipc}
\end{table}

\section{Data Augmentation Ablation}

To understand the contribution of data augmentation in our framework, we conduct a comparative study between two settings: standard image-based augmentations and our proposed Temporally Guided Video Augmentation. The results clearly show that traditional image augmentations tend to disrupt temporal coherence, resulting in suboptimal performance. In contrast, our TSGF$_A$ significantly enhances the temporal diversity of synthetic videos while preserving motion continuity, leading to noticeable performance gains.

\section{Experiments on Static and Dynamic Group}
Following the protocol established by VDSD\cite{wang2024dancing}, the MiniUCF dataset was partitioned into two subsets: the static group, consisting of categories characterized by minimal motion changes, and the dynamic group, comprising categories with significant temporal variations. We conducted separate distillation experiments on these two groups to investigate the effectiveness of our method under different temporal dynamics. Our results show that on the dynamic group, which requires capturing complex temporal dependencies, our method achieves outstanding performance. This demonstrates the superior capability of our framework in preserving and leveraging temporal dynamics during the distillation process. The distinction between static and dynamic groups highlights the importance of explicitly modeling temporal information in video dataset distillation. While static actions rely more on spatial cues, dynamic actions demand effective temporal modeling, which our method addresses through temporal saliency-guided optimization and augmentation.

\section{Visualization}

\subsection{Optical Flows}
\begin{wrapfigure}{r}{0.5\linewidth}
    \centering
    \includegraphics[width=0.9\linewidth]{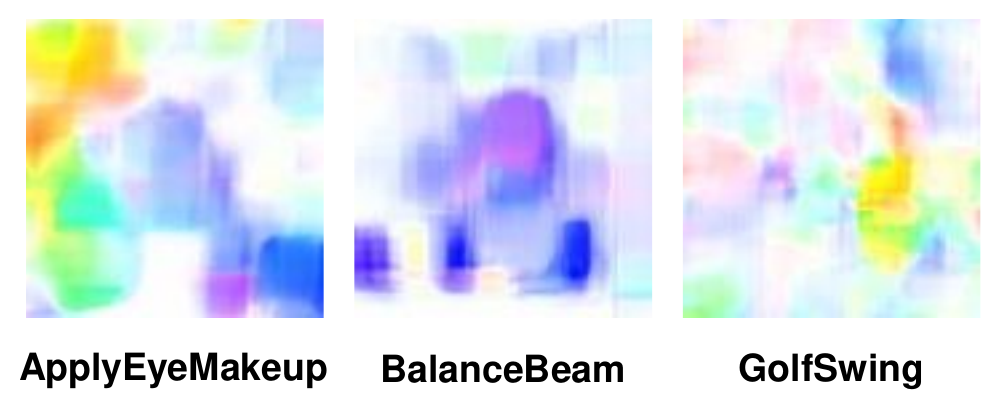}
    \caption{Optical flows of MiniUCF.}
    \label{fig:flow}
\end{wrapfigure}

To qualitatively analyze how our method preserves temporal dynamics, we visualize the optical flow of distilled videos in Figure \ref{fig:flow}. The optical flow maps clearly illustrate that our approach preserves smooth and consistent motion patterns. This indicates that the temporal dynamics crucial for action recognition are effectively retained during the distillation process.

\subsection{Inter-frame Differences}
\begin{figure}
    \centering
    \includegraphics[width=0.95\linewidth]{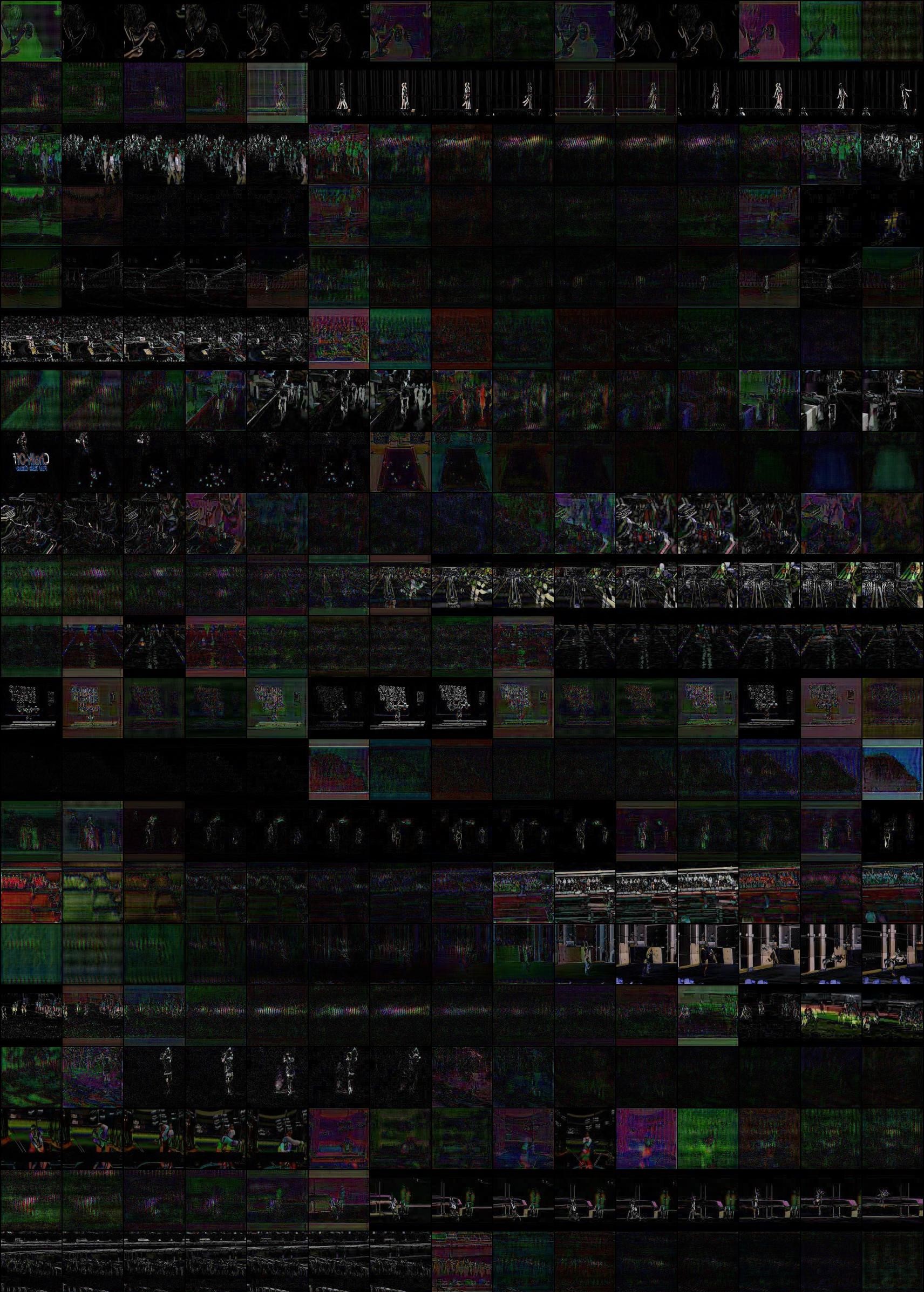}
    \caption{Inter-frame difference visualization on MiniUCF.}
    \label{fig:diff}
\end{figure}

In addition to optical flow, we also visualize frame differences to highlight temporal changes across frames. As shown in Figure \ref{fig:diff}, our method maintains coherent motion patterns and smooth temporal transitions. This further validates the effectiveness of our temporal saliency-guided framework in maintaining critical dynamic information necessary for accurate video understanding.

\section{Implementation Details}

\subsection{Hyperparameters}
\begin{wraptable}{r}{0.4\linewidth}
    \centering
    \resizebox{0.95\linewidth}{!}{
        \begin{tabular}{cc|cc}
            \toprule
            Dataset & IPC & lr & r\_bn \\
            \midrule
            \multirow{2}{*}{MiniUCF} & 1 & 0.25 & 0.001 \\ 
            & 5 & 0.25 & 0.005 \\
            \multirow{2}{*}{HMDB51} & 1 & 0.25 & 0.001 \\ 
            & 5 & 0.25 & 0.005 \\
            \multirow{2}{*}{Kinetics-400} & 1 & 0.3 & 0.01 \\ 
            & 5 & 0.3 & 0.01 \\
            \multirow{2}{*}{SSv2} & 1 & 0.3 & 0.01 \\ 
            & 5 & 0.3 & 0.01 \\
            \bottomrule
        \end{tabular}}
    \caption{Hyperparameters for different datasets.}
    \label{tab:hyper}
\end{wraptable}

We provide a detailed summary of the parameters used in our experiments in Table \ref{tab:hyper}, where lr denotes the learning rate, and r\_bn refers to the coefficient of the regularization loss. Parameters not explicitly stated are assumed to follow the default values specified in our code.

\subsection{Computational Resources}
All experiments were performed on a server equipped with eight NVIDIA RTX 3090 GPUs, each with 24 GB of memory. Compared to training on the full-scale original datasets, our video dataset distillation method significantly reduces both training time and computational resource demands.

\section{Broader Impact}
Our proposed video dataset distillation framework holds significant potential to advance research and applications in video understanding by substantially reducing the computational cost associated with training deep video models. This can democratize access to large-scale video analysis techniques, enabling wider adoption across academia and industry, especially for groups with limited computational resources. By facilitating faster and more efficient model training, our method may accelerate developments in areas such as video surveillance, autonomous driving, human-computer interaction, and content recommendation systems, ultimately benefiting society through improved safety, convenience, and personalized experiences. Importantly, the distilled synthetic datasets generated by our approach inherently protect privacy by reducing reliance on storing and sharing large volumes of original video data, which often contain sensitive or personally identifiable information. From an ethical perspective, our work does not introduce new risks for malicious use, as it focuses on improving data efficiency rather than altering or generating misleading content. Nonetheless, as with any technology related to video analysis, responsible deployment should consider privacy, consent, and potential biases in training data. Overall, we believe our method contributes positively to sustainable and ethical AI development in the video domain.

\end{document}